\documentclass[runningheads]{llncs}

\usepackage{eccv}

\usepackage{booktabs}
\usepackage{multirow}
\usepackage{makecell}
\usepackage{xcolor,colortbl}

\usepackage{eccvabbrv}

\usepackage{graphicx}
\usepackage{booktabs}

\usepackage[accsupp]{axessibility}  

\usepackage[utf8]{inputenc}

\usepackage{tcolorbox}
\tcbuselibrary{listingsutf8, breakable}

\newtcolorbox{promptbox}[2][]{breakable,
  colback=blue!5!white, colframe=blue!75!black,
  fonttitle=\bfseries, title=Prompt,
  label={#2}, #1}

\newtcolorbox{replybox}[2][]{breakable,
  colback=green!5!white, colframe=green!75!black,
  fonttitle=\bfseries, title=Model Response,
  label={#2}, #1}

\usepackage[pagebackref,breaklinks,colorlinks,citecolor=eccvblue]{hyperref}

\usepackage{orcidlink}

\begin{document}

\title{MeDocVL: A Visual Language Model for Medical Document Understanding and Parsing}

\titlerunning{MeDocVL}

\author{
Ping An Property \& Casualty Insurance Company
    \thanks{
    Preview: this manuscript is a preprint and may be updated.
    \newline Code: \url{https://github.com/Dejavuvvw/MeDocVL}.
    \newline Contact: \email{diaoliang145,zhaoyuan041@pingan.com.cn}.
    }
}
\authorrunning{VCG}

\institute{
Visual Computing Group (VCG), Shenzhen, China
}

\maketitle

\begin{abstract}
Medical document OCR is challenging due to complex layouts, domain-specific terminology, and noisy annotations, while requiring strict field-level exact matching.
Existing OCR systems and general-purpose vision–language models often fail to reliably parse such documents.
We propose \textbf{MeDocVL}, a post-trained vision–language model for query-driven medical document parsing.
Our framework combines \emph{Training-driven Label Refinement} to construct high-quality supervision from noisy annotations, with a \emph{Noise-aware Hybrid Post-training} strategy that integrates reinforcement learning and supervised fine-tuning to achieve robust and precise extraction.
Experiments on medical invoice benchmarks show that MeDocVL consistently outperforms conventional OCR systems and strong VLM baselines, achieving state-of-the-art performance under noisy supervision.
\keywords{OCR \and Vision--Language Model \and Medical Document Parsing \and Post-training}
\end{abstract}

\section{Introduction}
\label{sec:intro}

\begin{figure}[tb]
  \centering
  \includegraphics[width=0.9\textwidth]{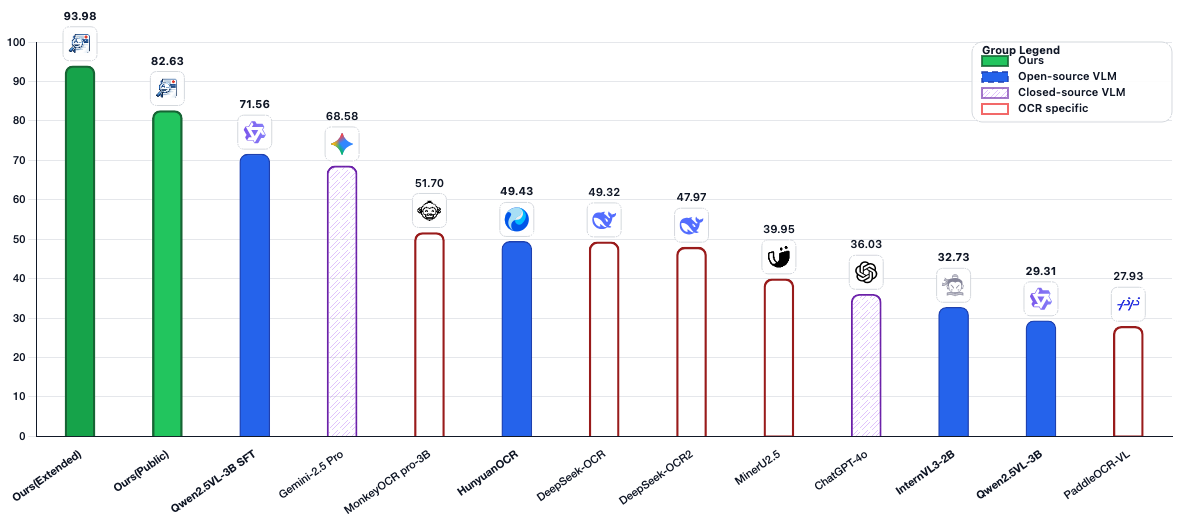}
  \caption{Comparison of query-driven document parsing performance.
  Public model is trained exclusively on publicly available datasets, while Extended model is trained with additional non-public data to study the effect of increased training data scale.}
  \label{fig:comparison_bar}
\end{figure}

\begin{figure}[tb]
  \centering
  \includegraphics[width=0.9\textwidth]{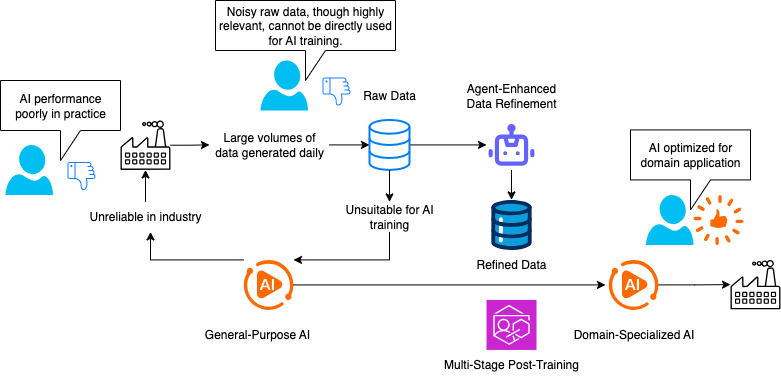}
  \caption{Overall pipeline of the proposed MeDocVL framework.
  Large-scale raw industrial data are highly relevant but unsuitable for direct training due to noisy or missing annotations.
  MeDocVL employs Training-driven Label Refinement to convert raw data into reliable domain supervision, followed by Noise-aware Hybrid Post-training to adapt a general-purpose VLM into a high-precision, domain-specialized model.}
  \label{fig:overall_pipeline}
\end{figure}

Modern Optical Character Recognition (OCR) has long been a cornerstone of information technology (IT) and continues to play an important role in promoting document digitization and information automation. 
Recent advances in artificial intelligence (AI) have significantly improved OCR performance and expanded its scope to document parsing, information extraction, and visual question answering. 
However, high-stakes, domain-specific document parsing remains challenging. 
In medical documents, insurance claims, and financial invoices, the objective is not merely transcription but accurate extraction of fields and their values (e.g., dates, amounts, identifiers), where a single-character error may invalidate the entire field and break downstream usage.

Meanwhile, practical document parsing systems increasingly require flexible, query-driven access to document content, allowing users to retrieve information via natural-language queries rather than a predefined schema. 
This motivates query-driven document content parsing, which demands both precise visual recognition and alignment with user intent.

Existing approaches rarely satisfy both requirements. 
Pipeline-based OCR systems convert images into intermediate representations (e.g., text/markup) and then rely on downstream components (often LLMs) for field extraction, which introduces error propagation, system complexity, and limited domain adaptability. 
In contrast, MLLM-based OCR/parsing models offer unified vision-language reasoning and natural-language interaction, but they remain unreliable for high-precision extraction, frequently producing token-level recognition errors (and occasional hallucinations) under dense layouts and strict formatting constraints.


These limitations reveal a fundamental gap, as high-precision visual recognition and flexible, query-driven parsing are rarely achieved simultaneously. 
We therefore study high-precision, end-to-end, query-driven document content parsing, focusing on medical documents and invoices. 
Our goal is to transform a general-purpose MLLM into a queryable document parsing expert that achieves OCR-level recognition fidelity while preserving the flexibility of natural-language querying. 
Crucially, we target realistic training conditions, where large-scale document data are available but annotations are incomplete or noisy—an unavoidable setting in industrial applications.

To this end, we propose a backbone-agnostic post-training framework for query-driven document content parsing under noisy supervision, built upon three tightly coupled technical components.
(1) \textbf{Training-driven Label Refinement (TLR)} formulates annotation correction as a learnable process. 
Starting from a small Clean Data with expert annotations, we generate structured pseudo-label prompts using OCR and MLLM systems, whose outputs constitute Noisy annotations. 
An Annotation Refinement Model is then trained to map these noisy annotations to corrected predictions, using expert labels only to supervise the refinement behavior. 
Importantly, TLR does not aim to eliminate all noise.
Instead, it distills the error characteristics of OCR/MLLMs and suppresses their systematic biases, producing Refined Data that provides more stable and reliable supervision for downstream optimization.
(2) Building upon the refined supervision, we introduce \textbf{Noise-aware Hybrid Post-training (NHP)}, which integrates supervised fine-tuning (SFT) and reinforcement learning (RL) in a noise-aware manner. 
Rather than directly fitting potentially unreliable annotations, RL is first employed on Refined Data with task-specific rewards to improve robustness under residual noise. 
Subsequently, SFT on Clean Data is used to consolidate extraction behavior, output formatting, and domain specialization. 
This staged design balances robustness and stability when adapting general-purpose MLLMs to domain-specific document parsing.
(3) Finally, we design a \textbf{token-wise GRPO objective for noise-robust document extraction}, which aligns RL with the exact-match requirements of structured field parsing. 
Instead of sequence-level optimization, we derive fine-grained token-level rewards from refined annotations, combined with token masking, optional confidence weighting, and reference regularization. 
This formulation focuses learning on reliable tokens and mitigates the adverse impact of noisy or ambiguous supervision.


The proposed framework is model-agnostic and applicable to a wide range of multimodal backbones. 
\textbf{MeDocVL} represents one concrete instantiation, enabling end-to-end, query-driven document parsing with recognition accuracy comparable to strong OCR baselines. 
Through systematic experiments, we analyze the impact of annotation noise, validate each component of the framework, and demonstrate improved robustness under noisy supervision. 

Our contributions can be summarized as follows:
\begin{itemize}
    \item  We formulate query-driven, high-precision document content parsing as a distinct problem setting that bridges OCR-level recognition accuracy with the flexibility of MLLMs.
    \item We propose a backbone-agnostic post-training framework for adapting general MLLMs to document parsing under noisy supervision, built upon Training-driven Label Refinement, Noise-aware Hybrid Post-training, and token-wise alignment.
    \item We conduct systematic experiments, including controlled noise injection, comparative evaluations, and ablation studies, to validate the role of each technical component.
    \item In addition, we release a curated medical and invoice document content parsing benchmark with refined annotations and evaluation protocols to facilitate reproducible research.
\end{itemize}

\section{Related Work}
\label{sec:related work}
Traditional document parsing approaches~\cite{breezedeus2025pix2text, livathinos2025doclingefficientopensourcetoolkit, niu2025mineru25decoupledvisionlanguagemodel} typically decompose document understanding into a sequence of specialized sub-tasks, including layout analysis~\cite{huang2022layoutlmv3, zhao2024detrs, wang2024yolov10}, reading order prediction~\cite{wang2021layoutreader}, optical character recognition (OCR)~\cite{jaidedaiEasyOCR, li2022pp}, formula recognition~\cite{wang2024unimernet}, and table structure recognition~\cite{padhi2021tabular, he2021pingan, xia2024docgenome}, each handled by an independent model.
Such modular pipelines are effective across diverse document formats and allow fine-grained optimization for individual components.
However, errors introduced at early stages tend to propagate downstream, and the lack of holistic semantic modeling often necessitates extensive post-processing to obtain coherent, human-readable outputs, limiting robustness in complex real-world scenarios.

The emergence of large language models (LLMs)~\cite{hui2024qwen2, liu2024deepseek, touvron2023llama} has partially alleviated these issues by providing strong text understanding and reasoning capabilities.
Several works~\cite{kim2022ocr, blecher2023nougat, wei2024general, peng2022spts, liu2023spts, wan2024omniparser} incorporate LLMs to aggregate OCR outputs, interpret document content, or perform structured information extraction.
While these methods improve semantic consistency and flexibility, LLMs operate purely on textual inputs and lack access to spatial and visual context, which makes them sensitive to layout errors and limits their reliability on documents with complex structures.

Recent vision–language models (VLMs)~\cite{openai2024helloGPT4o, zhu2025internvl3, zeng2025glm, bai2025qwen2} address this limitation by jointly modeling visual and textual information through a unified architecture.
By coupling a vision encoder with a language model, VLMs can directly process document images and reason over layout, typography, and semantics in an end-to-end manner, achieving strong performance on general OCR and document understanding benchmarks.
Representative models, including the Monkey series~\cite{li2024monkey, liu2024textmonkey, li2025monkeyocr}, mPLUG-DocOwl2~\cite{hu2024mplug}, and olmOCR~\cite{poznanski2025olmocr}, demonstrate the effectiveness of large-scale VLM fine-tuning on document-centric corpora.
Despite these advances, most existing VLM-based approaches prioritize general-purpose versatility and benchmark coverage, rather than the strict precision, reliability, and robustness required in domain-specific, high-stakes applications.

In settings such as medical document recognition, where field-level exactness, interpretability, and error tolerance are critical, the above limitations become particularly pronounced.
This motivates our proposed \textbf{MeDocVL} framework, which integrates training-driven data refinement with noise-aware post-training into a unified adaptation pipeline.
By explicitly addressing annotation noise, systematic OCR errors, and preference alignment during post-training, MeDocVL provides a scalable and robust approach for adapting general-purpose VLMs to specialized industrial domains.

\section{Training-driven Label Refinement}
\label{sec:tlr}
In this section, we introduce Training-driven Label Refinement (TLR), a principled, data-centric post-training component designed to stabilize learning for query-driven document content parsing under noisy supervision.
Unlike conventional data cleaning pipelines that aim to produce perfectly accurate annotations, TLR explicitly targets the dominant sources of systematic annotation noise arising from OCR systems and MLLMs, and seeks to reduce their bias while preserving realistic supervision variability.

In real-world document understanding scenarios, large-scale training data are typically obtained through a combination of human annotation and automated labeling systems.
While extensive filtering is often applied to remove severely corrupted samples, residual noise in annotations remains inevitable due to human mistakes and consistent failure patterns of OCR and MLLM-based extractors.
Such noise is particularly harmful for document content parsing, where supervision is defined at the field level, and even minor token-level errors can invalidate an entire key-value pair.
Directly applying SFT on noisy annotations therefore leads to unstable optimization and brittle extraction behavior.

TLR addresses this challenge by formulating label refinement itself as a learnable task.
Starting from a small Expert Annotations Dataset with reliable ground truth, we construct structured pseudo labels by jointly leveraging OCR systems and MLLMs.
These pseudo labels, although imperfect, encode rich domain knowledge and characteristic error patterns.
An Annotation Refinement Model is then trained to revise these pseudo labels by explicitly learning how OCR\/MLLM predictions deviate from expert annotations.
Importantly, this process does not aim to eliminate all noise; instead, it distills correction behavior and suppresses systematic biases, producing a Refined Data distribution that is more suitable for downstream optimization under noisy supervision.

The TLR pipeline consists of three stages:
\begin{enumerate}
    \item \textbf{Pseudo-label construction and prompt synthesis.}
    We generate structured pseudo labels from expert-annotated samples using OCR and MLLM predictions, and convert them into instruction-style prompts that expose typical annotation errors to the refinement model.
    \item \textbf{Correction distillation.}
    The Annotation Refinement Model is trained by comparing its corrected outputs against expert annotations, learning to revise noisy key--value pairs while preserving valid information from pseudo labels.
    \item \textbf{Large-scale refinement and filtering.}
    The trained refiner is applied to the Noisy Annotations Dataset to produce Refined Data, which exhibits reduced systematic noise and improved consistency, and serves as the primary supervision source for subsequent noise-aware post-training.
\end{enumerate}

By transforming noisy annotations into a more stable and structured supervision signal, TLR establishes a robust foundation for the Noise-aware Hybrid Post-training strategy and token-wise reinforcement learning objectives introduced in later sections.

\subsection{Stage 1: Pseudo-label Construction and Prompt Synthesis}

\begin{figure}[tb]
  \centering
  \includegraphics[width=0.9\textwidth]{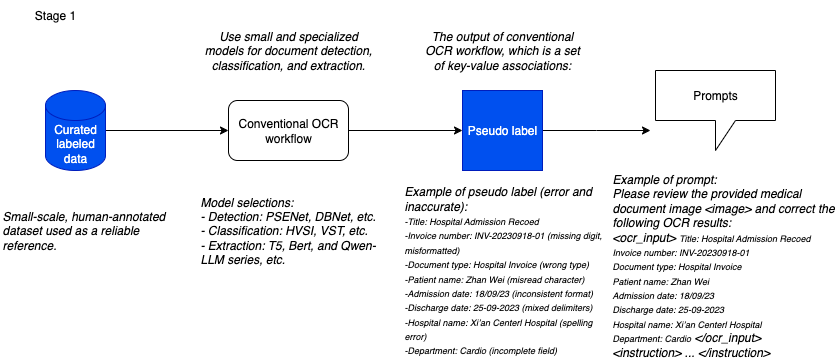}
  \caption{\textbf{Stage 1 — Pseudo-label construction and prompt synthesis.}
  Expert-annotated documents are processed by OCR systems and MLLMs to produce structured but imperfect key--value predictions.
  These predictions are converted into instruction-style prompts that explicitly expose annotation errors, enabling the refinement model to learn correction behaviors rather than generating labels from scratch.}
  \label{fig:data refinement stage1}
\end{figure}

The first stage of TLR focuses on constructing informative pseudo labels and corresponding instruction prompts that expose typical annotation errors to the refinement model.
Starting from a small Expert Annotations Dataset with reliable ground truth, we generate pseudo labels by jointly leveraging OCR systems and multimodal large language models (MLLMs).
These pseudo labels consist of structured key-value pairs that reflect realistic extraction outputs, including both correct fields and systematic errors commonly observed in medical documents.

Unlike conventional data augmentation or masking-based self-regression strategies, we intentionally preserve the full key-value structure of pseudo labels, even when values are incorrect.
This design choice is motivated by two observations.
First, OCR and MLLM predictions encode substantial domain knowledge and layout priors, which would be discarded if values were masked or regenerated from scratch.
Second, the dominant failure modes in document parsing arise from consistent, structured errors (e.g., misaligned fields, truncated spans, or semantic confusion between similar medical terms), which must be explicitly observed in order to be corrected.

To this end, pseudo labels are transformed into instruction-style prompts that present the model with both the document image and the predicted key-value pairs, and request the correction of erroneous fields.
The refinement model is therefore trained to revise existing predictions rather than hallucinating new ones.
This prompt formulation shifts the learning objective from unconditional generation to conditional correction, enabling the model to focus on identifying and fixing systematic deviations between pseudo labels and expert annotations.

Crucially, this process should not be interpreted as an attempt to construct noise-free supervision.
Instead, pseudo labels serve as a structured intermediate representation that captures the characteristic error distribution of upstream OCR and MLLM systems.
By conditioning on these imperfect predictions, the refinement model learns how annotation noise manifests at the field and token levels, which forms the foundation for subsequent correction distillation and large-scale refinement.

Through this design, Stage~1 establishes a supervision interface that maximizes the reuse of existing extraction signals, exposes realistic error patterns, and prepares the refinement model for learning robust correction behaviors in the presence of noisy annotations.

\subsection{Stage 2: Correction Distillation Training}

\begin{figure}[tb]
  \centering
  \includegraphics[width=0.9\textwidth]{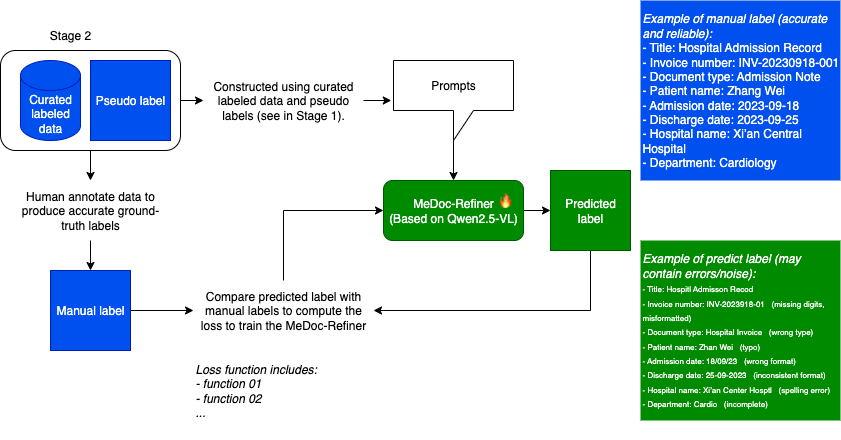}
  \caption{\textbf{Stage 2 — Correction distillation training.}
  The Annotation Refinement Model is trained to revise pseudo labels by comparing its predictions against expert annotations.
  The training objective emphasizes learning systematic correction behaviors while preserving the general multimodal representations of the base model.}
  \label{fig:data refinement stage2}
\end{figure}

Given the pseudo-label prompts constructed in Stage~1, the second stage focuses on training the Annotation Refinement Model to acquire robust correction capabilities.
The model is initialized from a pre-trained multimodal backbone (Qwen2.5-VL) and adapted using parameter-efficient fine-tuning.
Rather than performing broad re-training, this stage explicitly targets the ability to identify and correct systematic annotation errors introduced by OCR systems and MLLMs.

The key objective of correction distillation is not to directly predict ground-truth labels from scratch, but to learn how pseudo labels deviate from expert annotations.
For each curated sample, the refinement model is presented with the document image and the corresponding pseudo-label prompt, and is trained to produce a corrected version that aligns with expert-provided annotations.
This formulation encourages the model to focus on error localization and revision, instead of unconditional generation, which is known to be unstable under noisy supervision.

We formulate this training process as a structured distillation problem.
Pseudo labels serve as a teacher signal that encodes domain-specific priors and layout awareness, while expert annotations provide the target correction direction.
By optimizing the refinement model to bridge the gap between the two, we effectively distill the correction behavior that maps noisy but informative supervision to more reliable annotations.
Importantly, this process suppresses systematic bias without collapsing supervision diversity, resulting in refined labels that remain realistic for downstream training.

To support heterogeneous document structures, the correction objective integrates complementary supervision signals at different granularities, including field-level classification, sequence correction for text spans, and spatial alignment for tabular content.
In addition, regularization terms are introduced to preserve the general multimodal reasoning ability of the base model and prevent overfitting to the small expert-annotated set.
The full mathematical formulation of the optimization objective is provided in the Appendix.

Through iterative correction distillation, the Annotation Refinement Model learns stable and transferable error-correction strategies.
As demonstrated in our experiments, this stage plays a critical role in reducing annotation noise and establishing a reliable supervision foundation for large-scale refinement and subsequent noise-aware post-training.

The full mathematical formulation of the correction distillation objective, including all loss terms, notation, and optimization details, is provided in Appendix~\ref{app:correction_distillation}.

\subsection{Stage 3: Large-scale Refinement for Noise-aware Post-training}

\begin{figure}[tb]
  \centering
  \includegraphics[width=0.9\textwidth]{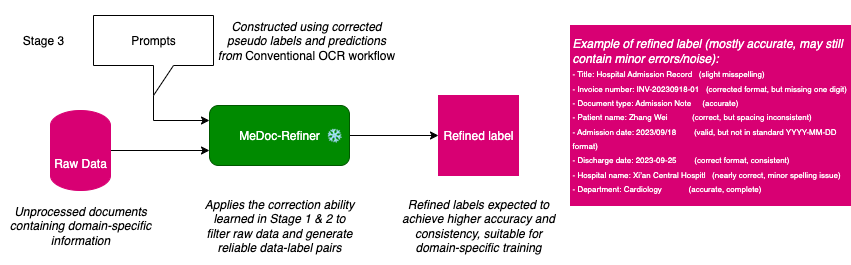}
  \caption{\textbf{Stage 3 — Large-scale refinement for noise-aware post-training.}
  The trained Annotation Refinement Model is applied to large-scale raw documents to transform noisy annotations into refined supervision.
  Rather than eliminating all errors, the resulting Refined Data reduces systematic bias while preserving realistic annotation variability, serving as a stable supervision source for noise-aware post-training.}
  \label{fig:data refinement stage3}
\end{figure}

In the final stage of TLR, the trained Annotation Refinement Model is deployed to process large volumes of domain-specific documents and generate Refined Data for downstream optimization.
Following the same prompt construction strategy as in Stage~1, the refiner revises OCR/MLLM-derived annotations by applying the correction behaviors learned during distillation.
The resulting annotations exhibit substantially reduced systematic errors compared to raw OCR outputs, while still retaining residual noise that reflects realistic extraction uncertainty.

Importantly, the objective of this stage is not to construct a perfectly clean dataset.
In practical industrial settings, large-scale document corpora inevitably contain annotation noise due to human mistakes and consistent failure patterns of automated labeling systems.
Over-aggressive filtering or manual correction may reduce noise but often leads to limited data scale and distorted supervision distributions.
Instead, TLR deliberately produces Refined Data that strikes a balance between consistency and diversity: systematic biases are suppressed, but controlled noise is preserved.

This property is crucial for subsequent Noise-aware Hybrid Post-training.
As shown in our experiments, supervised fine-tuning is highly sensitive to annotation noise and tends to overfit to corrupted labels, whereas reinforcement learning objectives exhibit greater tolerance to residual noise.
By reshaping the supervision distribution rather than fully denoising it, Refined Data provides a stable foundation on which noise-aware optimization strategies can effectively operate.

\begin{figure}[tb]
  \centering
  \includegraphics[width=0.5\textwidth]{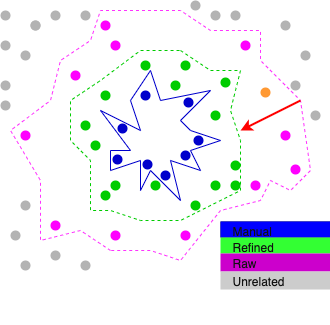}
  \caption{\textbf{Optimization landscape under different supervision sources.}
  Raw OCR annotations induce a broad and irregular optimization space that hampers convergence.
  Expert annotations define a narrow but potentially overfitting-prone region.
  Refined Data occupies an intermediate regime, reducing irregularity while maintaining sufficient coverage for stable and robust optimization.}
  \label{fig:optimization convergence}
\end{figure}

Figure~\ref{fig:optimization convergence} provides an intuitive illustration of this effect.
Raw OCR supervision yields a highly irregular optimization landscape, making learning unstable.
Expert annotations constrain optimization to a narrow region, which accelerates convergence but increases the risk of overfitting under limited data.
In contrast, Refined Data reshapes the supervision distribution into a smoother yet sufficiently diverse space, enabling robust optimization under noisy supervision.

Through this large-scale refinement process, TLR produces a supervision source that is cost-effective, scalable, and explicitly designed to support noise-aware post-training.
This stage bridges data refinement and optimization, and directly motivates the hybrid training strategy introduced in the next section.

\section{Noise-aware Hybrid Post-training}
\label{sec:train}

In this section, we introduce Noise-aware Hybrid Post-training (NHP), a post-training strategy designed to adapt multimodal large language models to high-precision document content parsing under noisy supervision.
Unlike conventional fine-tuning pipelines that rely solely on supervised learning, NHP explicitly accounts for the intrinsic noise and bias present in large-scale document annotations, and leverages complementary optimization paradigms to achieve both robustness and precision.

As established in Section~\ref{sec:tlr}, even after training-driven label refinement, large-scale supervision inevitably contains residual noise.
For document parsing tasks, this noise is particularly harmful, as supervision is defined at the field and token levels, and small annotation errors can lead to incorrect extraction outcomes.
Directly applying supervised fine-tuning (SFT) under such conditions often results in overfitting to corrupted labels and unstable optimization, especially when exact matching is required.

To address this challenge, NHP adopts a hybrid post-training paradigm that combines reinforcement learning (RL) and supervised fine-tuning in a noise-aware manner.
The core insight is that these two optimization regimes play fundamentally different roles under noisy supervision.
Reinforcement learning introduces soft preference constraints, allowing the model to improve alignment while tolerating imperfect annotations, whereas supervised fine-tuning enforces hard token-level targets, which is effective only once the model behavior has been sufficiently stabilized.

\begin{figure}[tb] 
    \centering 
    \includegraphics[width=0.9\textwidth]{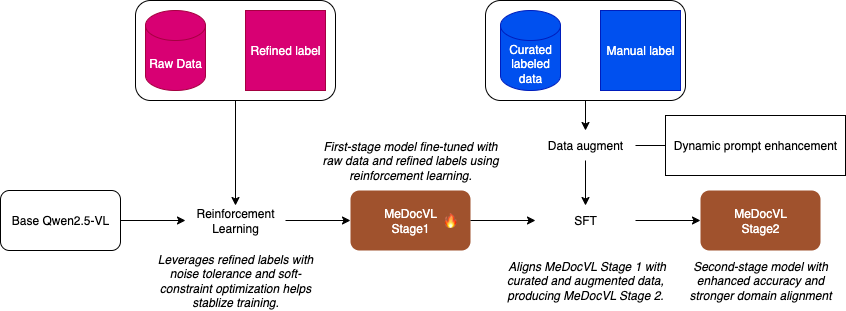} 
    \caption{Training recipe of MeDocVL. In Stage~1, raw data and initial labels are corrected by a VLM (\emph{MeDocVL-Stage1}) to generate refined data and labels. In Stage~2, the refined data and labels are used to further train a correction model (\emph{MeDocVL-Stage2}), enhancing the quality of data–label pairs and improving model robustness.} 
    \label{fig:training_recipe} 
\end{figure}

As illustrated in Figure~\ref{fig:training_recipe}, NHP decomposes post-training into two stages.
In the first stage, we apply reinforcement learning on Refined Data produced by TLR, using preference-based objectives to guide the model away from systematic errors without strictly enforcing noisy labels.
This stage emphasizes robustness and correction behavior under uncertainty.
In the second stage, we apply supervised fine-tuning on curated high-quality annotations to consolidate extraction accuracy and strengthen domain-specific behavior.
Importantly, the order of these stages is non-interchangeable: RL serves to stabilize learning under noise, while SFT refines and solidifies the aligned behavior.

Throughout NHP, the vision encoder of the multimodal backbone is kept frozen to preserve strong visual representations learned during pre-training.
By decoupling robustness-oriented alignment from precision-oriented refinement, NHP provides a principled and efficient post-training strategy for high-precision, query-driven document parsing in realistic noisy supervision settings.

\subsection{Reinforcement Learning under Noisy Supervision}
After Training-driven Label Refinement (TLR), the resulting supervision is substantially cleaner than raw OCR outputs, yet still inevitably contains residual noise.
For document content parsing, such noise is particularly harmful, as training signals are defined at the field and token levels, and even minor annotation errors can lead to incorrect extraction behavior.
Directly applying supervised fine-tuning (SFT) on this data risks overfitting to noisy labels and may push the model toward brittle solutions.

Motivated by recent findings that naïve SFT on multimodal reasoning data can degrade alignment performance~\cite{chen2025sft}, we adopt reinforcement learning (RL) as the first post-training stage.
Instead of enforcing strict one-to-one correspondence between predictions and labels, RL introduces \emph{preference-based soft constraints}, allowing the model to improve correction behavior while remaining tolerant to imperfect supervision.
This property makes RL particularly suitable for learning from Refiner-labeled data, where annotation noise is reduced but not eliminated.

\paragraph{Why reinforcement learning before SFT.}
Applying SFT prior to preference optimization has been shown to induce pseudo reasoning paths, i.e., superficial patterns that mimic correct outputs without learning robust correction strategies.
Once such patterns are reinforced, subsequent optimization becomes unstable.
In contrast, RL optimizes relative preferences between candidate outputs, guiding the model toward more reliable behaviors without over-committing to noisy targets.
This stage therefore serves as a behavioral stabilization step before high-precision supervision is enforced in later SFT.

\paragraph{From sequence-level to token-level preference optimization.}
We implement a customized variant of Generalized Reinforcement Preference Optimization (GRPO).
Standard GRPO computes token-level rewards but aggregates them into sequence-level advantages, while GSPO further averages these rewards to improve stability for long-chain reasoning tasks.
However, document recognition fundamentally differs from reasoning-centric tasks: correctness depends on precise token predictions (e.g., digits, dates, medical codes), rather than global sequence coherence.
Aggregating rewards at the sequence level may obscure localized errors that are critical for document parsing.

To address this mismatch, we adopt token-wise preference optimization, where preference signals are applied directly at the token level.
This design provides fine-grained corrective supervision, ensuring that small but critical extraction errors are explicitly penalized.
The resulting optimization objective emphasizes local accuracy while retaining the robustness advantages of RL.

\subsubsection{Token-wise GRPO Objective}
Given an input document $x$ with a preferred response $y^{+}$ and a dispreferred response $y^{-}$, we define the token-wise preference loss as:
\begin{equation}
\label{eq:tok-grpo}
\mathcal{L}_{\mathrm{tok\text{-}GRPO}}
=
\mathbb{E}_{(x,y^{+},y^{-})}
\left[
-\frac{1}{T}\sum_{t=1}^{T}
w_t \log \sigma\!\Big(
\beta \big[\Delta_t(\theta) - \kappa\,\Delta^{\mathrm{ref}}_t\big]
\Big)
\right],
\end{equation}
where $\Delta_t(\theta)$ denotes the token-level log-probability gap between preferred and dispreferred responses, $w_t$ is a token confidence weight, and $\Delta^{\mathrm{ref}}_t$ constrains policy drift relative to the frozen backbone.
Full definitions and derivations are provided in Appendix~\ref{app:tok_grpo}.

\paragraph{Stabilization via KL regularization.}
To preserve the strong general representations of the pre-trained backbone, we include a KL regularization term that limits deviation from the reference policy.
The detailed formulation is provided in Appendix~\ref{app:rl_kl}.
The overall RL objective is:
\begin{equation}
\label{eq:rl-total}
\mathcal{L}_{\mathrm{RL}}
=
\mathcal{L}_{\mathrm{tok\text{-}GRPO}}
+
\lambda_{\mathrm{KL}}\,\mathcal{L}_{\mathrm{KL}}.
\end{equation}

\paragraph{Discussion.}
Compared to sequence-level GRPO or GSPO, token-wise GRPO better aligns with the error characteristics of document recognition.
By weighting tokens according to confidence and avoiding unnecessary length penalties, the optimization remains both stable and precise.
This RL stage equips the model with robust correction behavior under noisy supervision, forming a reliable foundation for subsequent supervised fine-tuning on clean data.

\subsection{Supervised Fine-tuning on Clean Data (Precision Consolidation Stage)}
\label{subsec:sft}

Following reinforcement learning under noisy supervision, we perform a final stage of Supervised Fine-Tuning (SFT) on clean, high-confidence annotations.
Unlike the preceding RL phase, which primarily focuses on robustness and preference alignment under imperfect supervision, this stage serves as a precision consolidation step.
Its goal is to sharpen token-level accuracy, reduce residual variance, and calibrate the model’s predictions to satisfy the strict exact-match requirements of document content parsing.

Document recognition differs fundamentally from open-ended reasoning tasks in that correctness is defined by fine-grained, field-level precision rather than long-chain inference.
After RL has guided the model toward robust and noise-tolerant behaviors, SFT provides strong deterministic supervision to consolidate these behaviors into stable and highly accurate extraction patterns.
Importantly, this stage operates exclusively on clean or refined annotations produced by the Training-driven Label Refinement (TLR) pipeline, avoiding the risk of reintroducing noisy supervision.

To mitigate catastrophic forgetting and maintain balanced performance across heterogeneous document formats, we adopt a mixed-stage SFT strategy.
Instead of sequentially fine-tuning the model on individual domains, all data sources are jointly used in a single SFT phase.
This design preserves cross-domain generalization while allowing the model to specialize in high-precision document parsing.

\subsubsection{Instruction Data Construction}
The SFT dataset is constructed from curated and refined instruction–response pairs, covering approximately 2 million samples.
The data are evenly distributed across three representative document categories: free-form documents, tabular documents, and mixed-format documents.
Each sample follows a unified instruction-driven format, explicitly specifying target fields to be extracted from the input document.

Both text-only and multimodal samples are included to maintain alignment with the pre-trained language capabilities of the backbone.
However, multimodal samples are prioritized for domain-specific consolidation, as they expose the model to realistic visual–textual interactions involving layout, tables, stamps, and handwritten content.
This balanced construction enables MeDocVL to consolidate high-precision extraction skills without sacrificing general instruction-following behavior.

\subsubsection{Dynamic Prompt-based Precision Augmentation}
To further strengthen precision and robustness, we introduce dynamic prompt-based augmentation during SFT.
Instead of perturbing the document images, controlled noise is injected into the instruction prompts.
Specifically, individual fields within the prompt are randomly masked or replaced with plausible but incorrect values, forcing the model to verify information against the visual content rather than relying on prompt priors.

Each field is assigned a perturbation probability in the range $[0,1]$, and fields with probabilities exceeding a threshold of $0.3$ are modified.
This threshold is selected empirically and validated in Section~\ref{sec:experiments}.
The perturbation process is lightweight yet effective, as it encourages systematic cross-checking between textual instructions and visual evidence.

This strategy directly complements the preceding RL phase.
While RL improves tolerance to annotation noise and preference misalignment, dynamic prompt augmentation during SFT tightens decision boundaries at the token level.
As a result, the model becomes less prone to hallucinated field values and more reliable in exact-match extraction scenarios.

Overall, this precision consolidation stage transforms the robustness gained from reinforcement learning into stable, high-accuracy document parsing performance, completing the Noise-aware Hybrid Post-training (NHP) pipeline.

\section{Experiments}
\label{sec:experiments}
We conduct extensive experiments to evaluate the effectiveness of our proposed method for query-driven document content parsing.
Specifically, we compare MeDocVL with state-of-the-art (SOTA) OCR and MLLM-based approaches for medical document parsing, analyze the impact of annotation noise on model fine-tuning, assess the effectiveness of training-driven label refinement, and perform ablation studies to verify the contribution of each component in our framework.

\subsection{Experiment Settings}

\subsubsection{Dataset preparation}
\begin{figure}[tb]
  \centering
  \includegraphics[width=0.9\textwidth]{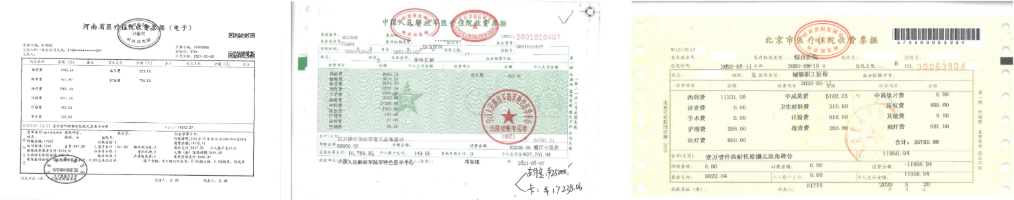}
  \caption{Representative samples from the medical invoice dataset.
  Although all documents belong to a single category (medical invoices), they exhibit substantial visual and semantic complexity, including domain-specific medical terminology, mixed layout structures, printed text, tabular regions, handwritten annotations, and official seals, posing significant challenges for accurate recognition and field-level extraction.}
  \label{fig:dataset_samples}
\end{figure}

We evaluate our method on a publicly available medical invoice dataset released by Alibaba Cloud~\cite{tianchi2023medicalocr}.
The dataset contains 800 real-world medical billing documents collected from diverse scenarios, with 700 images used for training and 100 for testing.

To better reflect practical application requirements, we re-annotated the dataset by refining and expanding field definitions while preserving the original document content and fixed data split.
Semantically redundant fields are consolidated, and additional fine-grained fields commonly required in real-world medical invoice processing are introduced.
The same re-annotated labels and experimental settings are applied to all baselines and our method, ensuring a fair and controlled comparison with full reproducibility.

Although the dataset contains a single document category, medical invoices exhibit substantial intra-class variability and domain-specific complexity.
As illustrated in Figure~\ref{fig:dataset_samples}, each document integrates heterogeneous information sources, including invoice metadata, medical terminology, itemized billing records, and patient-related fields, and often combines printed text, tabular layouts, handwritten content, and official stamps or seals.
These characteristics result in dense layouts and complex visual–textual alignments, posing significant challenges for accurate recognition and query-driven field extraction.

\subsubsection{Metrics}
Our experiments aim to assess practical field-level recognition performance under realistic document parsing scenarios.
Conventional sequence-level metrics, such as exact match accuracy, are often insufficient to capture extraction quality across heterogeneous fields and may lack discriminative power when documents contain multiple keys.

We therefore adopt Field Match Ratio (FMR) as the primary evaluation metric.
Since document annotations are represented as key–value pairs, FMR measures the proportion of correctly extracted field values across documents, directly reflecting field-level correctness.

Formally, let $D$ denote the set of documents, $K_d$ the set of keys in document $d$, $v_{d,k}$ the ground-truth value for key $k$, and $\hat{v}_{d,k}$ the predicted value.
Since labels are represented as key–value associations, the model must accurately extract the correct value for each key. 
We define the micro-averaged FMR as:
\begin{equation}
\mathrm{FMR}_{\mathrm{micro}}
= \frac{\sum_{d\in D}\sum_{k\in K_d} \left[\phi(\hat v_{d,k})=\phi(v_{d,k})\right]}
       {\sum_{d\in D} |K_d|},
\end{equation}
where $\phi(\cdot)$ denotes optional normalization (e.g., case folding or date reformatting).
For completeness, we also report the macro-averaged FMR across keys:
\begin{equation}
\mathrm{FMR}_{\mathrm{macro}}
= \frac{1}{|K|}\sum_{k\in K}
  \left(
    \frac{\sum_{d\in D_k} \left[\phi(\hat v_{d,k})=\phi(v_{d,k})\right]}{|D_k|}
  \right),
\end{equation}
where $K=\bigcup_{d\in D} K_d$ and $D_k=\{d\in D \mid k\in K_d\}$.
Unless otherwise specified, we report micro-averaged FMR in all experiments.

\subsection{Comparison with the SOTA Models}

\begin{table*}[!t]
    \begin{center}
    \caption{FMR $\uparrow$ on medical invoice}
    \label{table:comparison_experiment}
    \small
    \setlength{\tabcolsep}{5pt}
        \begin{tabular}{c|c}
            \hline
            \textbf{Method} & \textbf{Medical Invoice} \\
            \hline
            \hline
            Qwen2.5-VL 3B       & 0.2931 \\
            InternalVL3 2B      & 0.5995 \\
            \hline
            \hline
            DeepSeek-OCR        & 0.4932 \\
            PaddleOCR-VL        & 0.2793 \\
            MonkeyOCR Pro 3B    & 0.517 \\
            MinerU2.5	        & 0.3995 \\ 
            \hline
            \hline
            ChatGPT-4o	        & 0.3603 \\ 
            Gemini-2.5 Pro	    & 0.6858 \\ 
            \hline
            \hline
            MeDoc-VL (Proprietary) & \textbf{0.9398} \\
            MeDoc-VL (Public) & \textbf{0.8244} \\
            \hline
        \end{tabular}
    \par\smallskip
    {\footnotesize Note: FMR = Field Match Ratio, best results per column are in \textbf{bold}.}
    \end{center}
\end{table*}

Table~\ref{table:comparison_experiment} reports the Field Match Ratio (FMR) of different methods on the medical invoice dataset.
We compare MeDoc-VL with representative OCR systems and MLLMs for document understanding.

The compared methods include general-purpose MLLMs with native visual–textual reasoning capabilities, such as Qwen2.5-VL (3B), InternalVL3 (2B), ChatGPT-4o, and Gemini-2.5 Pro, as well as OCR-oriented or document-focused models, including DeepSeek-OCR, PaddleOCR-VL, MonkeyOCR Pro (3B), and MinerU2.5.
Notably, most OCR-based models do not natively support query-driven document parsing, i.e., they cannot directly retrieve fields or values conditioned on user-provided natural-language queries.

To enable a fair comparison under a unified query-driven setting, we adopt a two-stage evaluation protocol for such methods.
Specifically, OCR models are first used to extract document content into structured JSON representations.
A fixed Qwen3-8B model is then employed as an intermediate parser to extract fields and values from the OCR outputs based on user queries.
The same extraction prompt and Qwen3 configuration are used across all OCR-based baselines, and the prompt is provided in the appendix to ensure reproducibility.
All methods are ultimately evaluated under the same query-driven document parsing protocol and measured using FMR.

As shown in Table~\ref{table:comparison_experiment}, MeDoc-VL achieves the highest FMR on medical invoices, substantially outperforming both OCR-based pipelines and general-purpose MLLMs.
While MLLMs such as Qwen2.5-VL and Gemini-2.5 Pro demonstrate reasonable performance, they still exhibit noticeable field-level errors under complex medical invoice layouts.
OCR-centric approaches, even when combined with a strong downstream parser, remain limited by recognition errors and imperfect semantic alignment, resulting in lower FMR.

We report two configurations of MeDoc-VL, denoted as Public and Extended, which differ only in the source of supervision used during post-training.
The Extended model is trained with additional large-scale proprietary data and is included solely to demonstrate the scalability and upper-bound performance of the proposed framework.
Due to privacy and compliance constraints, its weights are not publicly released.
The Public model is trained exclusively on the publicly available Alibaba Cloud dataset and is fully reproducible.
Both configurations achieve identical performance on this benchmark, indicating that the proposed framework is robust to supervision sources and does not rely on proprietary data.

\subsection{Noisy Effectiveness Analysis}
\begin{table}[t]
    \centering
    \small
    \caption{Impact of annotation noise on supervised fine-tuning and the effectiveness of training-driven label refinement.}
    \begin{tabular}{cc}
        \hline
        \textbf{Training Data} & \textbf{FMR} \\
        \hline
        20\% Noisy Data & 0.7812 \\
        30\% Noisy Data & 0.7296 \\
        50\% Noisy Data & 0.5820 \\
        70\% Noisy Data & 0.2601 \\
        Refined Data (from 30\% noise) & \textbf{0.8133} \\
        \hline
    \end{tabular}
    \label{tab:noise_analysis}
\end{table}

This experiment investigates the impact of annotation noise on SFT and evaluates the effectiveness of Training-driven Label Refinement (TLR) in mitigating noise-induced performance degradation.
Starting from the same foundation model, we synthetically inject noise into training annotations by randomly corrupting a controlled proportion of key-value labels, with noise ratios set to 20\%, 30\%, 50\%, and 70\%.

We emphasize that noise is injected into annotations rather than raw document images to faithfully simulate practical post-training scenarios.
In real-world document processing pipelines, input data are typically pre-filtered to remove severely corrupted or low-quality samples, ensuring the visual content itself is relatively reliable.
However, annotations often contain non-negligible noise due to human labeling errors or systematic biases introduced by automatic annotation systems (e.g., OCR or MLLM-based labeling).
Accordingly, we focus on modeling annotation noise rather than data corruption.

To reflect realistic annotation errors, corrupted labels are not replaced with random characters.
Instead, values are substituted with semantically plausible but incorrect alternatives (e.g., valid-looking dates, amounts, or medical terms), preserving surface-level realism while violating field-level correctness.
All models are then fine-tuned under identical settings and evaluated on the clean test set.

As shown in Table~\ref{tab:noise_analysis}, model performance degrades rapidly as the noise ratio increases.
Even moderate noise (30\%) leads to a noticeable drop in FMR, while higher noise levels (50\% and 70\%) cause a severe performance collapse.
These results indicate that SFT for document content parsing is highly sensitive to annotation noise, primarily due to the strict field-level exact-match requirements.

We further evaluate the effectiveness of label refinement by applying the Annotation Refinement Model to the 30\% Noisy Data, producing a Refined Data set.
Fine-tuning on Refined Data yields an FMR of 0.8133, which not only recovers the performance loss caused by noisy supervision but also outperforms all SFT results trained on unrefined noisy annotations.
This demonstrates that TLR effectively suppresses systematic annotation noise and provides more reliable supervision for downstream optimization.

Overall, this analysis confirms that annotation noise constitutes a critical bottleneck for SFT in query-driven document parsing, and that training-driven label refinement serves as an effective mechanism for stabilizing learning under realistic noisy supervision.

\subsection{Ablation Study}
\begin{table*}[!t]
    \begin{center}
    \caption{Ablation study results under different model configurations.}
    \small
    \setlength{\tabcolsep}{5pt}
    \label{table:ablation_study}
        \begin{tabular}{l|l|c}
            \toprule
            \multicolumn{2}{c|}{\textbf{Pipeline Configuration}} & \textbf{Invoice} \\
            \cmidrule(lr){1-2}\cmidrule(lr){3-3}
            \textbf{Stage-1} & \textbf{Stage-2} & FMR  \\
            
            \midrule
            \rowcolor[HTML]{F7F7F7}
            \multicolumn{3}{l}{\textbf{(A) Baseline}} \\
            \midrule
            \multicolumn{2}{l|}{\textbf{baseline}} & 0.2931 \\
            
            \midrule
            \rowcolor[HTML]{F7F7F7}
            \multicolumn{3}{l}{\textbf{(B) Single-Stage Training}} \\
            \midrule
            \textbf{SFT on clean data}      & \textemdash & 0.753  \\
            \textbf{SFT on corrected data}  & \textemdash & 0.8133 \\
            \textbf{Vanilla RL}             & \textemdash & 0.8197 \\
            \textbf{Token-wise GRPO}        & \textemdash & 0.8214 \\
            \midrule
            
            \rowcolor[HTML]{F7F7F7}
            \multicolumn{3}{l}{\textbf{(C) Two-Stage Training}} \\
            \midrule
            \textbf{SFT}                & \textbf{SFT}          & 0.8152 \\
            \textbf{SFT}                & \textbf{Vanilla RL}   & 0.8198 \\
            \textbf{Vanilla RL}         & \textbf{SFT}          & 0.8225 \\
            \textbf{Token-wise GRPO}    & \textbf{SFT}          & 0.8244 \\
            \midrule
            
            \rowcolor[HTML]{F7F7F7}
            \multicolumn{3}{l}{\textbf{(D) Full Pipeline}} \\
            \midrule
            \textbf{Token-wise GRPO} & \makecell[l]{\textbf{SFT} \\+ \textbf{DynPrompt}} & \textbf{0.8263} \\
            
            \bottomrule
        \end{tabular}
    \par\smallskip
    {\footnotesize Best results per column are in \textbf{bold}.}
    \end{center}
\end{table*}

We conduct comprehensive ablation studies to analyze the contribution of each component in the proposed framework and to compare the behavior of SFT and RL under noisy supervision.
All variants are evaluated on the medical invoice dataset using the same evaluation protocol, and results are summarized in Table~\ref{table:ablation_study}.
Unless otherwise stated, all ablation results are reported using the Public MeDocVL setting.

\paragraph{Baseline and Single-Stage Training.}
The baseline model achieves an FMR of 0.2931, indicating that direct foundation model under noisy supervision yields limited performance.
As shown in Block (B), single-stage SFT on clean data provides only marginal improvement, while SFT on corrected data leads to a substantial gain, demonstrating the importance of refined supervision.
Single-stage RL further improves performance, with token-wise GRPO outperforming vanilla RL, highlighting the benefit of fine-grained reward design even without multi-stage training.

\paragraph{RL vs. SFT under Noisy Supervision.}
Block (C) compares different two-stage training orders.
Notably, configurations that apply RL before SFT consistently outperform those that apply SFT before RL.
This observation suggests that SFT is more prone to overfitting noisy or partially corrected annotations, whereas RL better tolerates residual noise by optimizing task-level rewards.
Applying SFT after RL helps consolidate extraction behavior and formatting, resulting in improved overall performance.
These results empirically confirm that RL exhibits higher robustness to noisy supervision than SFT in query-driven document parsing.

\paragraph{Effect of Token-wise GRPO.}
Across both single-stage and two-stage settings, token-wise GRPO consistently outperforms vanilla RL.
This improvement can be attributed to its token-level optimization objective, which directly targets field-level correctness and avoids diluting localized extraction errors that are common in sequence-level reinforcement learning.

\paragraph{Full Pipeline.}
The full pipeline, which integrates token-wise GRPO followed by SFT with dynamic prompt augmentation, achieves the best performance with an FMR of 0.8534.
Compared with the baseline, this represents a significant improvement, validating the complementary roles of training-driven label refinement, noise-aware hybrid post-training.
Overall, the ablation results demonstrate that each component contributes to performance gains and that their combination is essential for robust, high-precision document content parsing under realistic noisy supervision.

\section{Conclusion}
We presented \textbf{MeDocVL}, a noise-aware post-training framework for high-precision medical document parsing with vision–language models.
By integrating Training-driven Label Refinement and Noise-aware Hybrid Post-training, our approach effectively addresses annotation noise, complex layouts, and strict field-level accuracy requirements.
Experimental results demonstrate that MeDocVL consistently outperforms conventional OCR systems and strong general-purpose VLMs.

Beyond medical documents, the proposed framework offers a scalable and cost-effective pathway for adapting VLMs to other high-stakes vertical domains.
Moreover, the refined supervision and specialist models produced by our pipeline can serve as reliable data sources for downstream domain specialization and post-training distillation in general-purpose multimodal models.

\section*{Contributors}

\begin{itemize}
  \item \textbf{Project Leaders:} 
  Zengjian Fan, Wenfeng Xie, Ziling Lin, De Shi, Lin Huang, Kaihe Xu, Hong Li
  \item \textbf{Core Contributors:} 
  Wenjie Wang, Wei Wu, Ying Liu, Yuan Zhao, Xiaole Lv, Liang Diao
\end{itemize}

\newpage

\newpage
\appendix

\section{Correction Distillation Objective}
\label{app:correction_distillation}

\subsection{Notation}
We define the notation used in the correction distillation stage as follows:
\begin{itemize}
    \item $x$: model input, consisting of a document image and a structured instruction prompt.
    \item $y = (y_1, \dots, y_T)$: expert-annotated target token sequence.
    \item $\hat{y} = (\hat{y}_1, \dots, \hat{y}_T)$: predicted token sequence produced by the refinement model.
    \item $p_{\theta}(\cdot \mid x)$: token distribution of the refinement model with parameters $\theta$.
    \item $p_{T}(\cdot \mid x)$: teacher distribution induced by pseudo labels (e.g., OCR/MLLM predictions).
    \item $\theta_0$: frozen parameters of the pre-trained base model.
    \item $\mathcal{D}_{\text{exp}}$: expert-annotated dataset.
    \item $\mathcal{D}_{\text{gen}}$: optional replay buffer for representation preservation.
    \item $\tau$: temperature parameter for distillation.
\end{itemize}

\subsection{Token-level Correction Distillation}

To align the refinement model with informative pseudo-label distributions while correcting their deviations, we adopt a token-level knowledge distillation objective.
Let $p_T^{(\tau)}$ and $p_\theta^{(\tau)}$ denote the softened teacher and student distributions with temperature $\tau$.
The distillation loss is defined as:
\begin{equation}
\mathcal{L}_{\text{KD}}
= \frac{1}{|\mathcal{D}_{\text{exp}}|}
\sum_{(x,y)\in\mathcal{D}_{\text{exp}}}
\frac{1}{T}
\sum_{t=1}^{T}
\tau^2 \,
\mathrm{KL}\!\left(
p_T^{(\tau)}(\cdot \mid x, y_{<t})
\;\big\|\;
p_\theta^{(\tau)}(\cdot \mid x, y_{<t})
\right).
\end{equation}

To avoid propagating uncertain supervision, distillation is applied only to tokens where the teacher confidence exceeds a predefined threshold.

\subsection{Task-Specific Supervision}

To support heterogeneous document structures, we integrate complementary task-specific objectives.

\paragraph{Field-level Classification.}
For document or field type supervision, we use a cross-entropy loss:
\begin{equation}
\mathcal{L}_{\text{CLS}}
= -\mathbb{E}_{(x,c)\in\mathcal{D}_{\text{exp}}}
\log p_\theta(c \mid x).
\end{equation}

\paragraph{Sequence Correction.}
For text spans extracted from documents, we apply a sequence-level negative log-likelihood loss:
\begin{equation}
\mathcal{L}_{\text{SEQ}}
= -\mathbb{E}_{(x,y)\in\mathcal{D}_{\text{exp}}}
\frac{1}{T}\sum_{t=1}^{T}
\log p_\theta(y_t \mid x, y_{<t}).
\end{equation}

\paragraph{Spatial Alignment.}
For structured elements such as tables, we incorporate a localization-aware alignment objective combining L1 distance, IoU loss, and classification loss over matched cells.

\subsection{Representation Preservation}

To prevent overfitting to the small expert-annotated set and preserve the general multimodal capabilities of the base model, we introduce regularization terms.

\paragraph{Parameter Regularization.}
\begin{equation}
\mathcal{L}_{\text{SP}}
= \sum_{p} w_p \left\| \theta_p - (\theta_0)_p \right\|_2^2.
\end{equation}

\paragraph{Logit Preservation.}
Optionally, a replay buffer $\mathcal{D}_{\text{gen}}$ is used to enforce logit-level consistency:
\begin{equation}
\mathcal{L}_{\text{KLP}}
= \mathbb{E}_{(x,y)\in\mathcal{D}_{\text{gen}}}
\frac{1}{T}\sum_{t=1}^{T}
\mathrm{KL}\!\left(
p_{\theta_0}(\cdot \mid x, y_{<t})
\;\big\|\;
p_{\theta}(\cdot \mid x, y_{<t})
\right).
\end{equation}

\subsection{Overall Objective}

The full correction distillation objective is a weighted combination of the above terms:
\begin{equation}
\mathcal{L}_{\text{TOTAL}}
= \alpha \mathcal{L}_{\text{KD}}
+ \beta \mathcal{L}_{\text{CLS}}
+ \gamma \mathcal{L}_{\text{SEQ}}
+ \delta \mathcal{L}_{\text{ALIGN}}
+ \varepsilon \left( \mathcal{L}_{\text{SP}} + \mathcal{L}_{\text{KLP}} \right),
\end{equation}
where $\alpha,\beta,\gamma,\delta,\varepsilon$ are hyperparameters.

\section{Details of Token-wise GRPO}
\label{app:tok_grpo}

This appendix provides the full formulation and technical details of the proposed token-wise Generalized Reinforcement Preference Optimization (GRPO), which is briefly introduced in Section~\ref{sec:train}.

\subsection{Preference Data Construction}

We construct a preference dataset
$\mathcal{D}=\{(x, y^{+}, y^{-})\}$,
where $x$ denotes the input document (image and instruction prompt),
$y^{+}$ is a preferred response, and
$y^{-}$ is a dispreferred response.

For document parsing, preferences are derived from:
(1) refined annotations produced by the MeDoc-Refiner,
(2) alternative candidate outputs generated by the model,
and (3) heuristic or rule-based validators (e.g., field consistency, format validity).
Unlike reasoning-centric tasks, preferences are typically localized to specific fields or spans rather than the entire sequence.

\subsection{Token-level Log-probability Gaps}

Let $y^{+}=(y^{+}_1,\dots,y^{+}_T)$ and $y^{-}=(y^{-}_1,\dots,y^{-}_T)$ denote preferred and dispreferred sequences.
Given model parameters $\theta$, we define the token-level log-probability gap as:
\begin{equation}
\Delta_t(\theta)
=
\log p_\theta\!\big(y^{+}_t \mid x, y^{+}_{<t}\big)
-
\log p_\theta\!\big(y^{-}_t \mid x, y^{-}_{<t}\big).
\end{equation}

To prevent excessive deviation from the pre-trained backbone, we additionally introduce a frozen reference policy $p_{\mathrm{ref}}$ (Qwen2.5-VL) and compute:
\begin{equation}
\Delta^{\mathrm{ref}}_t
=
\log p_{\mathrm{ref}}\!\big(y^{+}_t \mid x, y^{+}_{<t}\big)
-
\log p_{\mathrm{ref}}\!\big(y^{-}_t \mid x, y^{-}_{<t}\big).
\end{equation}

These quantities capture relative preference strength at each token position.

\subsection{Token-wise GRPO Objective}

The token-wise GRPO loss is defined as:
\begin{equation}
\mathcal{L}_{\mathrm{tok\text{-}GRPO}}
=
\mathbb{E}_{(x,y^{+},y^{-}) \sim \mathcal{D}}
\left[
-\frac{1}{T}\sum_{t=1}^{T}
w_t
\log \sigma\!\Big(
\beta\big[\Delta_t(\theta)-\kappa\,\Delta^{\mathrm{ref}}_t\big]
\Big)
\right],
\end{equation}
where:
\begin{itemize}
    \item $\sigma(\cdot)$ is the sigmoid function;
    \item $\beta>0$ controls preference sharpness;
    \item $\kappa\ge0$ regulates reference-policy correction;
    \item $w_t \in [0,1]$ is a token-level confidence weight.
\end{itemize}

The confidence weight $w_t$ allows selective learning:
tokens corresponding to unreliable OCR spans or ambiguous fields can be down-weighted or masked ($w_t=0$),
while high-confidence tokens receive stronger supervision.

\subsection{Relation to GRPO and GSPO}

Standard GRPO aggregates token-level rewards into sequence-level advantages, while GSPO further averages rewards to improve stability in long-chain reasoning.
For document recognition, this aggregation is suboptimal, as correctness depends on localized token accuracy rather than holistic sequence coherence.

Token-wise GRPO removes this aggregation step and applies preference optimization directly at the token level.
This design ensures that fine-grained extraction errors (e.g., single-digit mistakes) are explicitly penalized, which is critical for high-precision document parsing.
latex
Copy code
\section{KL Stabilization and Reference Policy}
\label{app:rl_kl}

To stabilize reinforcement learning and preserve the strong general representations of the pre-trained backbone, we incorporate KL regularization with respect to a frozen reference policy.

\subsection{Token-level KL Divergence}

Given a preferred response $y^{+}$ and a dispreferred response $y^{-}$, we define the token-averaged KL divergence as:
\begin{equation}
\begin{aligned}
\mathcal{L}_{\mathrm{KL}}
&=
\mathbb{E}_{(x,y^{+})}
\left[
\frac{1}{T}\sum_{t=1}^{T}
\mathrm{KL}\!\Big(
p_\theta(\cdot \mid x, y^{+}_{<t})
\;\|\;
p_{\mathrm{ref}}(\cdot \mid x, y^{+}_{<t})
\Big)
\right]
\\
&\quad+
\mathbb{E}_{(x,y^{-})}
\left[
\frac{1}{T}\sum_{t=1}^{T}
\mathrm{KL}\!\Big(
p_\theta(\cdot \mid x, y^{-}_{<t})
\;\|\;
p_{\mathrm{ref}}(\cdot \mid x, y^{-}_{<t})
\Big)
\right].
\end{aligned}
\end{equation}

This symmetric formulation constrains both preferred and dispreferred trajectories, preventing policy collapse or uncontrolled drift.

\subsection{Total Reinforcement Learning Objective}

The final RL objective combines token-wise preference optimization with KL regularization:
\begin{equation}
\mathcal{L}_{\mathrm{RL}}
=
\mathcal{L}_{\mathrm{tok\text{-}GRPO}}
+
\lambda_{\mathrm{KL}}\,\mathcal{L}_{\mathrm{KL}},
\qquad
\lambda_{\mathrm{KL}} \ge 0.
\end{equation}

The hyperparameter $\lambda_{\mathrm{KL}}$ balances stability and adaptability.
Larger values enforce conservative updates, while smaller values accelerate domain adaptation.
In all experiments, $\lambda_{\mathrm{KL}}$ is selected to ensure stable convergence without sacrificing correction accuracy.

\subsection{Discussion}

KL stabilization plays a critical role in adapting large multimodal models to domain-specific document parsing tasks.
By anchoring optimization to the pre-trained backbone, the model retains general visual–textual representations while gradually acquiring robust correction behaviors under noisy supervision.

\bibliographystyle{splncs04}
\bibliography{main}
\end{document}